%% file: arxiv.tex
\documentclass[10pt,twocolumn,letterpaper]{article}

\usepackage[pagenumbers]{cvpr} 

\input{preamble}

\definecolor{cvprblue}{rgb}{0.21,0.49,0.74}
\usepackage{color, colortbl}
\definecolor{Gray}{gray}{0.95}
\definecolor{Gray_blue}{rgb}{0.92,0.92,1.0}
\usepackage[pagebackref,breaklinks,colorlinks,citecolor=cvprblue]{hyperref}

\def\paperID{13149} 
\def\confName{CVPR}
\def\confYear{2024}

\title{Rethinking Transformers Pre-training for Multi-Spectral Satellite Imagery}

\author{Mubashir Noman$^1$ \quad Muzammal Naseer$^1$ \quad Hisham Cholakkal$^1$ \quad Rao Muhammad Anwar$^{1}$ \\Salman Khan$^{1, 2}$ \quad Fahad Shahbaz Khan$^{1,3}$
\vspace{0.4em}
\\
$^1$Mohamed bin Zayed University of AI \quad $^2$Australian National University \\$^3$Linköping University
}

\begin{document}
\maketitle
\begin{abstract}
Recent advances in unsupervised learning have demonstrated the ability of large vision models to achieve promising results on downstream tasks by pre-training on large amount of unlabelled data. Such pre-training techniques have also been explored recently in the remote sensing domain due to the availability of large amount of unlabelled data. Different from standard natural image datasets, remote sensing data is acquired from various sensor technologies and exhibit diverse range of scale variations as well as modalities. Existing satellite image pre-training methods either ignore the scale information present in the remote sensing imagery or restrict themselves to use only a single type of data modality. In this paper, we re-visit transformers pre-training and leverage multi-scale information that is effectively utilized with multiple modalities. Our proposed approach, named SatMAE++, performs multi-scale pre-training and utilizes convolution based upsampling blocks to reconstruct the image at higher scales making it extensible to include more scales. Compared to existing works, the proposed SatMAE++ with multi-scale pre-training is equally effective for both optical as well as multi-spectral imagery. Extensive experiments on six datasets reveal the merits of proposed contributions, leading to state-of-the-art performance on all datasets. SatMAE++ achieves mean average precision (mAP) gain of 2.5\% for multi-label classification task on BigEarthNet dataset. Our code and pre-trained models are available at \url{https://github.com/techmn/satmae_pp}.

\end{abstract}

\section{Introduction}
Remote sensing employs a wide range of sensor technologies to acquire data for earth observation and monitoring through satellites and aircrafts. The acquired data may differ in terms of Ground Sample Distance (GSD) based on sensor technology and altitude. GSD refers to the distance between the two adjacent pixels measured on ground i.e., GSD of 0.3 meter in an image means adjacent pixels in the image are 0.3 meter apart on the ground. Accordingly, an image with size of $10 \times 10$ pixels may span over the city depending on the GSD of the image. Therefore, scale of the objects may vary considerably within a single image. In addition, multi-spectral (Sentinel-2) data captured from satellite utilize different sensors to acquire data. Consequently, multi-spectral data may possess different GSDs within the single image. Various bands combination in multi-spectral data can be used to highlight different information. For example, short wave and near infrared bands combination may be used for crop and agriculture monitoring. Therefore, it is desired to utilize the multi-scale information from various sensor's data for remote sensing tasks.

Even though the utilization of pre-trained models on large volume of data for remote sensing applications are increasingly popular, multi-scale information in remote sensing data is scarcely exploited by the vision community. Recently, self supervised learning on remote sensing data is widely explored \cite{ayush2022geographyaware, mañas2021seco, mendieta2023geospatial, neumann2019indomain}. SatMAE \cite{cong2023satmae} demonstrate the effectiveness of pre-training the transformers on large amount of data for various downstream remote sensing tasks. However, SatMAE \cite{cong2023satmae} does not exploit the multi-scale information present in the remote sensing satellite imagery and is not able to generalize over domains having images with multiple scales. ScaleMAE \cite{reed2023scalemae} is a newly introduced framework that proposes a strategy to encode the multi-scale information present in optical remote sensing data. The authors propose a GSD based positional encoding (GSDPE) to inform the model about the position and scale of the patch tokens. However, the proposed GSDPE is constrained to be utilized with the RGB (optical) images only. As multi-spectral (Sentinel-2) images have different GSD resolutions for different channels (see Tab.~\ref{tbl:sentinel2_desc} for more description) and ScaleMAE requires image channels to be stacked together \cite{reed2023scalemae}, therefore, GSDPE cannot be utilized with multi-spectral data.
Additionally, ScaleMAE introduced a sophisticated Laplacian pyramid based decoder enabling the model to learn multi-scale representations.
Alternatively, SatMAE shows that the standard sinusoidal position encodings can be easily extended for multi-spectral data. We argue that the MAE can still learn better multi-scale representations without using GSDPE and sophisticated Laplacian based decoder. To this end, we propose a framework SatMAE++ that uses the standard position encoding and reconstruct the image at multiple scale levels by utilizing convolution based upsampling blocks. In general, our contributions are:
\begin{itemize}
\item We empirically demonstrate that the standard position encodings along with multi-scale reconstruction encourage the model to learn better feature representations.
\item Based on our observations, we propose a simple yet effective multi-scale pre-training approach, named SatMAE++, that achieves impressive performance on both optical as well as multi-spectral satellite imagery.
\item We perform extensive experimentation on six datasets to validate the efficacy of our multi-scale pre-training framework. Our SATMAE++ sets a new state-of-the-art on all six datasets. On the downstream task of land cover classification, our SatMAE++ achieves an absolute gain of 3.6\% over the baseline \cite{cong2023satmae}. 
\end{itemize}

\begin{figure*}[t!]
   \centering
    \includegraphics[width=1.0\linewidth]{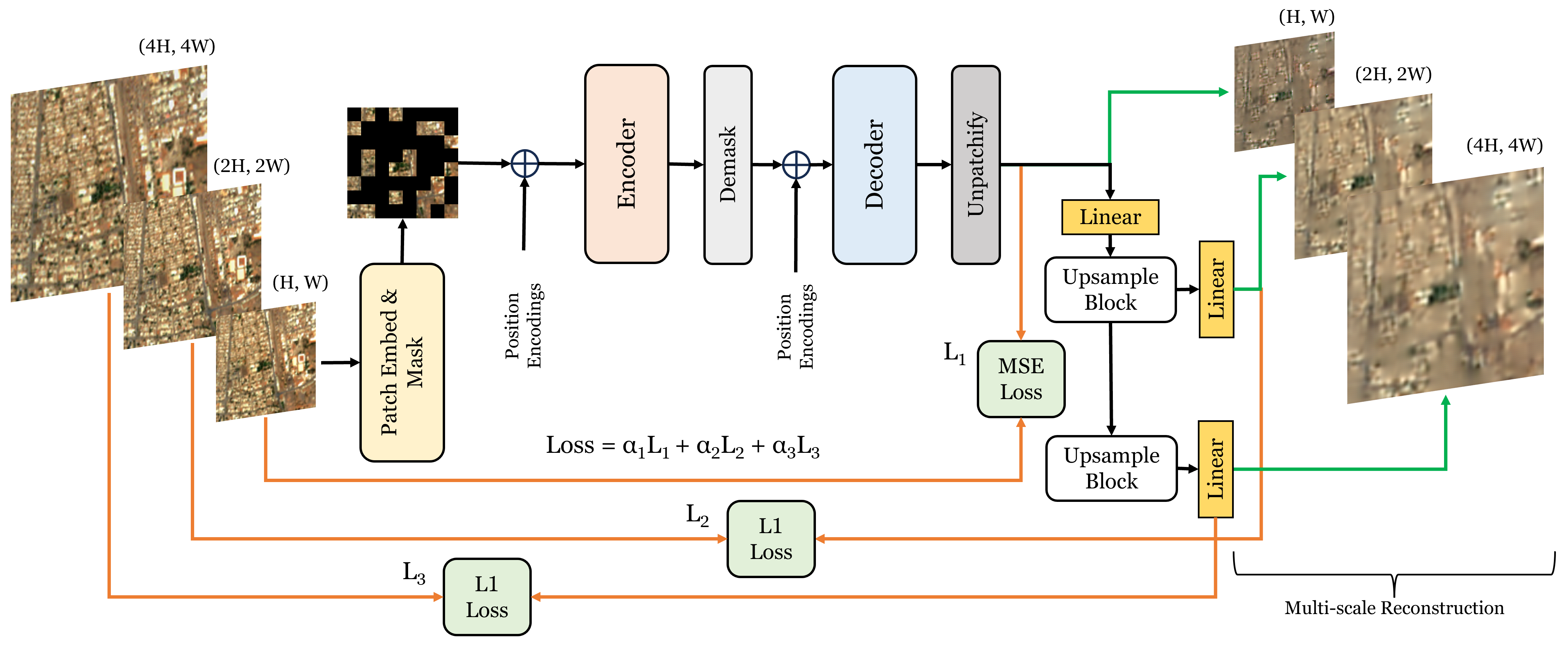}
    \caption{Illustration of Mask Autoencoder (MAE) framework for SatMAE++. The input image having spatial resolution of $(4H, 4W)$ is downsampled twice to obtain the images of resolution $(2H, 2W)$ and $(H, W)$, respectively. We then feed the image with resolution $(H, W)$ to the MAE similar to the SatMAE \cite{cong2023satmae} framework. The decoder reconstructs the image at the resolution $(H, W)$ and apply MSE loss to measure reconstruction quality. The reconstructed output is projected back to the feature space and is upsampled through upsampling blocks to obtain features at $(2H, 2W)$ and $(4H, 4W)$ resolutions. The upsampled outputs are projected back to the image space and L1 loss is utilized to penalize the reconstructions at higher resolutions. The overall loss is the weighted mean of all the losses. }    
   \label{fig:overall_architecture}
   \vspace{-0.8em}
\end{figure*}

\section{Related Work}
\subsection{Representation Learning for Satellite Imagery} Recently, self-supervised learning has been widely utilized as pre-training step for various remote sensing tasks enabling the model to learn rich feature representation from unlabelled data. GASSL \cite{ayush2022geographyaware} and SeCo \cite{mañas2021seco} employ the contrastive learning strategy to demonstrate the effectiveness of pre-training on different downstream tasks. Neumann et. al. \cite{neumann2019indomain} explore the idea of in-domain representation learning for remote sensing imagery. Mendieta et. al. \cite{mendieta2023geospatial} investigated the continual learning strategy and mask image modeling technique in their work. SatMAE \cite{cong2023satmae} utilized the mask image modeling for multi-spectral, temporal and optical remote sensing data. ScaleMAE \cite{reed2023scalemae} introduces a ground sample distance (GSD) based positional encoding for optical satellite imagery to encode the scale information present in the remote sensing data. While ScaleMAE has shown promising results on optical satellite data, its complicated GSD positional encoding limits its ability to be utilized for multi-spectral data. In contrast, we simplify the idea of extracting multi-scale information in optical and multi-spectral satellite data.

\subsection{Multi-scale Information} Visual images typically contain objects of various sizes and scales.
It is therefore desired to utilize the multi-scale information to learn better semantic representations. Convolutional neural networks and transformers \cite{liu2021swin, he2015resnet, elgcnet, yang2022focalnet, lin2017FPN, fan2021multiscalevit, noman2023remote, he2018maskrcnn, rssurveyaman2023} have employed the multi-scale information and show promising results on classification, detection, and segmentation tasks. ConvMAE \cite{gao2022convmae} introduces a hierarchical masking strategy to learn multi-scale features for a hybrid convolution-transformer encoder. Point-M2AE \cite{zhang2022pointm2ae} proposes a multi-scale auto encoder for 3D point clouds. ScaleMAE \cite{reed2023scalemae} propose a Laplacian based decoder to learn multi-scale information.
We rationalize the method and introduce an extensible convolution based upsampling block to reconstruct the feature maps at multiple scales.

\begin{table}[t]
\centering
\caption{Sentinel-2 bands description. The multi-spectral data contains three different GSD resolutions ranging from 10m to 60m. Different band combinations can be utilized to highlight specific information in the multi-spectral image.}
\vspace{-0.5em}
\label{tbl:sentinel2_desc}
\scalebox{0.75}{
\begin{tabular}{l|c|c|c}
\hline
\textbf{Band} & \textbf{Description} &\textbf{GSD (m)} & \textbf{Central Wavelength (nm)} \\
\hline
B1 & Ultra Blue (Aerosol) & 60 & 443 \\
B2 & Blue & 10 & 490 \\
B3 & Green & 10 & 560 \\
B4 & Red & 10 & 665 \\
B5 & Red Edge 1 & 20 & 705 \\
B6 & Red Edge 2 & 20 & 740 \\
B7 & Red Edge 3 & 20 & 783 \\
B8 & Near Infra Red & 10 & 842 \\
B8A & Red Edge 4 & 20 & 865 \\
B9 & Water Vapor & 60 & 940 \\
B10 & Cirrus & 60 & 1375 \\
B11 & Short Wave Infra Red 1 & 20 & 1610 \\
B12 & Short Wave Infra Red 2 & 20 & 2190 \\
\hline
\end{tabular}
}
\vspace{-1.0em}
\end{table}

\section{Background}\label{sec:vanilla_mae}
\textbf{Masked Auto-Encoder (MAE)} takes an input $I \in \mathbb{R} ^ {C \times H \times W}$, and resizes the image into sequence $S$ of non-overlapping patches having patch size $P$, where $S \in \mathbb{R} ^ {N \times P^2C}$ and $N = (\frac{H}{P}.\frac{W}{P})$ is the number of patches. A patch embedding layer $f: \mathbb{R}^{P^2C} \rightarrow \mathbb{R}^D$ is utilized to obtain the sequence $\bar S \in \mathbb{R}^{N \times D}$ of embedded tokens. A fraction $m$ of the $N$ tokens are masked and the remaining $1-m$ tokens are input to the transformer encoder that utilizes the position embedding to capture the spatial information of the patches in the image. In decoder, the visible tokens are placed back to their original sequence positions and learnable mask tokens are appended to obtain $N$ tokens. The positional embeddings are added to the $N$ tokens and fed to a series of transformer blocks. The decoder finally reconstruct the image by utilizing the mean squared error loss.

\noindent\textbf{Positional Encodings} Transformers utilize the positional encodings to learn the spatial locations of the patch tokens. In general, transformers use the sinusoidal positional encodings given as:

\begin{equation}
\label{eq:pos_encoding}
    \begin{aligned}
        v_x(pos, 2i) = sin(\frac{k}{\Omega^\frac{2i}{d}}) \\
        v_y(pos, 2i+1) = cos(\frac{k}{\Omega^\frac{2i}{d}}) \\
   \end{aligned}
\end{equation}

where $pos$ is the position, $i$ is feature dimension index, $d$ is the total possible positions, and $\Omega$ is a large constant.

\section{Method}
\subsection{Baseline Framework}
We adapt the recent SatMAE \cite{cong2023satmae} framework as our baseline. SatMAE follows the vanilla MAE based architecture for RGB data as discussed in section \ref{sec:vanilla_mae}. To deal with multi-spectral data, SatMAE performs channel grouping based on the GSDs of the multi-spectral channels. To make the training stable and reduce the memory requirements of the self attention operation, spatial size of the input is reduced to $96 \times 96$ instead of the standard input size of $224 \times 224$ \cite{cong2023satmae}. Accordingly, patch size is decreased to 8. Then it creates separate patch embeddings layers and concatenate them in the spatial dimension. SatMAE modifies the position encoding of MAE to incorporate the spectral information. To this end, separate encoding for each channel group is created and concatenated to the $x_{pos,i}, y_{pos,i}$ positional encodings such that the final encoding dimension is $D$. Afterwards, it passes the features to the transformer encoder. Similar to the vanilla MAE, the decoder takes the output of transformer encoder, place the visible tokens to their original position and learnable mask tokens are appended to obtain the $N$ tokens. Subsequently, the spectral encodings are added to the patch tokens and feed to the series of transformer blocks. Finally, mean squared error (MSE) loss is utilized to measure the reconstruction quality.

\noindent\textbf{Limitations: } Although the above baseline framework operates well on RGB and multi-spectral data, however it does not exploit the multi-scale information present in the remote sensing data. In addition, the above baseline framework strives to generalize across domains having images at multiple scale levels.
Furthermore, ScaleMAE \cite{reed2023scalemae} shows that the scale information can be encoded through a GSD based positional embeddings.
However, the GSD based positional embeddings introduced by ScaleMAE can only be used with the RGB data that has same GSD resolution for each channel \cite{reed2023scalemae}.
Therefore, it is desired to rethink the design of the SatMAE framework to learn the multi-scale information that is not constrained to single data modality.

\subsection{Overall Architecture}
Fig.~\ref{fig:overall_architecture} illustrates the overall framework of SatMAE++ that overcomes the limitations of baseline framework SatMAE \cite{cong2023satmae} and other recent approaches for multi-scale pre-training on multi-spectral (fMoW-Sentinel) as well as RGB data (fMoW-RGB). We take input image at most of three scale levels and feed the image at lowest scale level to the SatMAE framework. The base framework takes the input image, apply the patch embedding and masking operations, and feed the visible tokens to the transformer encoder. Subsequently, decoder takes the output of the encoder and reconstructs the image having same spatial dimension as the lowest scale level input. The reconstructed output from the SatMAE model is utilized by the upsample blocks to perform fine grained reconstruction at higher scale levels. Reconstruction at higher scales encourage the model to learn multi-scale representations thereby improving the performance on various downstream tasks.

\begin{figure}[t!]
   \centering
    \includegraphics[width=0.85\linewidth]{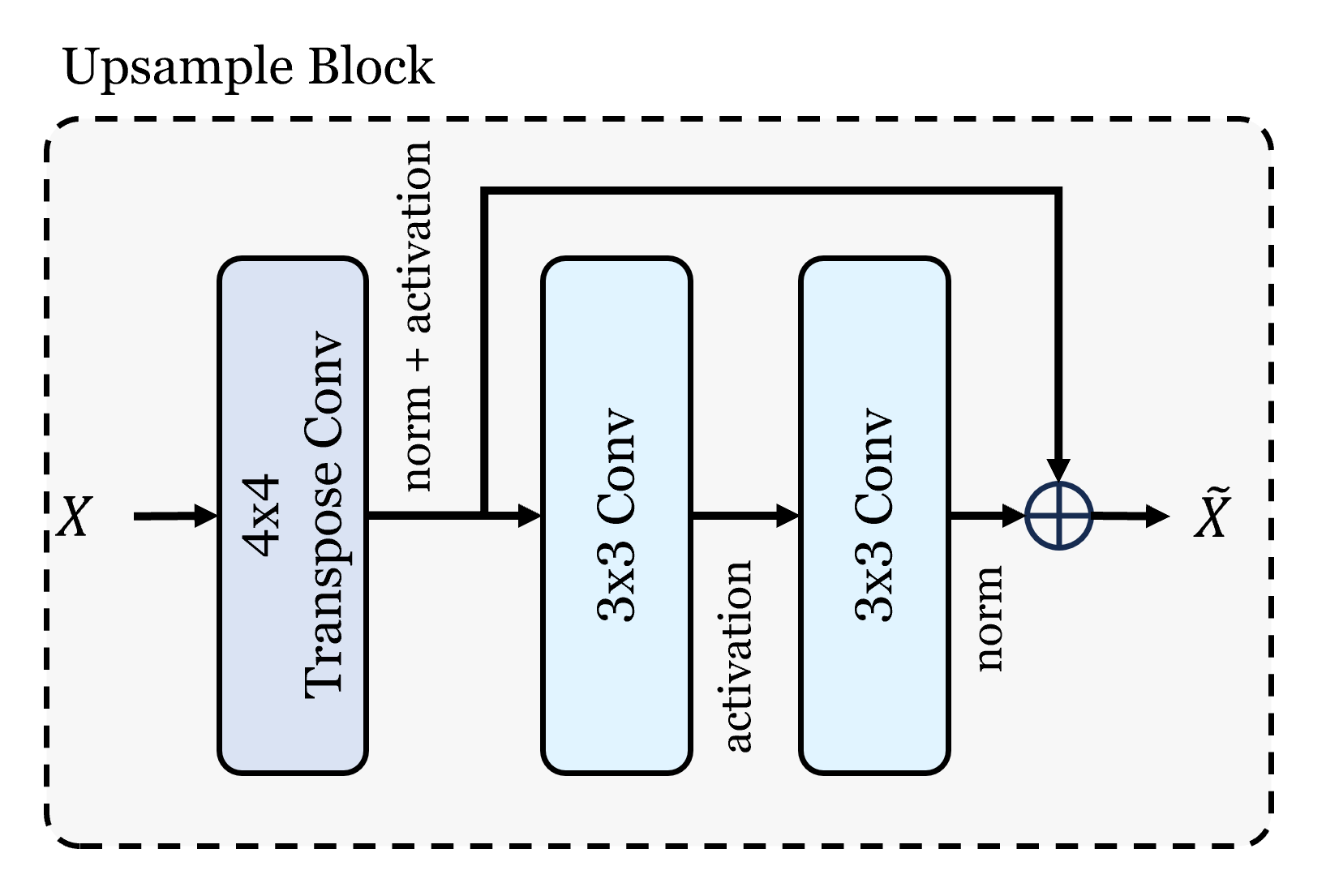}
    \caption{The illustration of upsample block used in SatMAE++ framework. Input features $X$ are upsampled by utilizing the transpose convolution operation. Afterwards, a residual block which is composed of two convolution layers is employed to enhance the upsampled features given as $\Tilde{X}$.}    
   \label{fig:upsample_block}
   \vspace{-1.0em}
\end{figure}

\noindent\textbf{Upsample Block: } Fig.~\ref{fig:upsample_block} shows the architecture of the upsample block. It takes input features $X \in \mathbb{R}^{C \times H \times W}$ and pass them through a transpose convolution layer to upsample the spatial resolution of the features. We normalize the upsampled features and apply the leaky relu activation operation afterwards. Then, a residual block, comprising of two convolution layers, is utilized for local feature enhancement. The enhanced features $\Tilde{X}  \in \mathbb{R}^{C \times 2H \times 2W}$ are projected back to the spatial domain by using a linear projection layer and mean absolute error is utilized to compute the reconstruction error between the input and the reconstructed image. Next, we discuss in detail the multi-scale reconstruction procedure at two and three scale levels.

\subsubsection{Reconstruction at Two Scale Levels}
Let $I \in \mathbb{R}^{C \times H \times W}$ be an input image to the MAE. We take an image $\hat I$ of resolution $\mathbb{R}^{C \times 2H \times 2W}$ and down-scale it to obtain the image $I$ of size $\mathbb{R}^{C \times H \times W}$.
The image $I$ is fed into the MAE which first utilizes the patch embedding layers to patchify the input image. In case of multi-spectral input, separate patch embedding layers is used for different groups of band channels. Afterwards, patch tokens of different groups are concatenated along the spatial dimension. We mask the 75\% of the patch tokens similar to the other MAE methods. Then, positional encodings are added to the visible patch tokens. 

Following \cite{cong2023satmae}, we utilize the general positional encodings (as illustrated in Eq.~\ref{eq:pos_encoding}) that does
not depend on the GSD information. The visible patch tokens are fed to a series of transformer blocks that produce the encoded visible features. Similar to the SatMAE, decoder takes the encoded visible features from the transformer encoder and apply a linear projection to reduce the embedding dimensions. Then, visible features are placed back to their original index positions and learnable mask tokens are appended to the visible tokens. Afterwards, RGB or multi-spectral positional encodings are added to the patch tokens. Eventually, the patch tokens are fed to the decoder transformer and a final projection layer maps the decoded features back to the spatial domain. The decoded image $F$ is reshaped to the original input dimensions and mean squared error is utilized to compute reconstruction quality.
\begin{equation}
    \label{eq:losses_two_scales_mse}
\resizebox{.4\hsize}{!}{
    $L_1 = \frac{1}{n}\sum_{i=1}^{n}{(F - I)}^2$
}
\end{equation}
After obtaining the reconstructed input from the decoder, we use linear projection to map the reconstructed image back to the feature space. We then utilize transpose convolution to upsample the feature maps at a resolution of $(2H \times 2W)$. The upsampled feature maps are passed from a residual block that is composed of two convolution layers.
Finally, we project the features back to the image space to obtain the scaled reconstructed image $\hat F$ and apply the L1 loss to analyze the reconstruction performance of model at a higher scale.
\begin{equation}
    \label{eq:losses_two_scales_mae}
\resizebox{.4\hsize}{!}{
    $L_2 = \frac{1}{n}\sum_{i=1}^{n}{|\hat F - \hat I|}$
}
\end{equation}
Likewise super-resolution methods, we utilize the L1 loss at higher scales so that the model can learn to reconstruct the actual image values. Overall loss is given as:
\begin{equation}
    \label{eq:losses_two_scales}
    Loss = \alpha_1 L_1+\alpha_2 L_2
\end{equation}

\begin{figure*}[t!]
   \centering
    \includegraphics[width=0.9\linewidth]{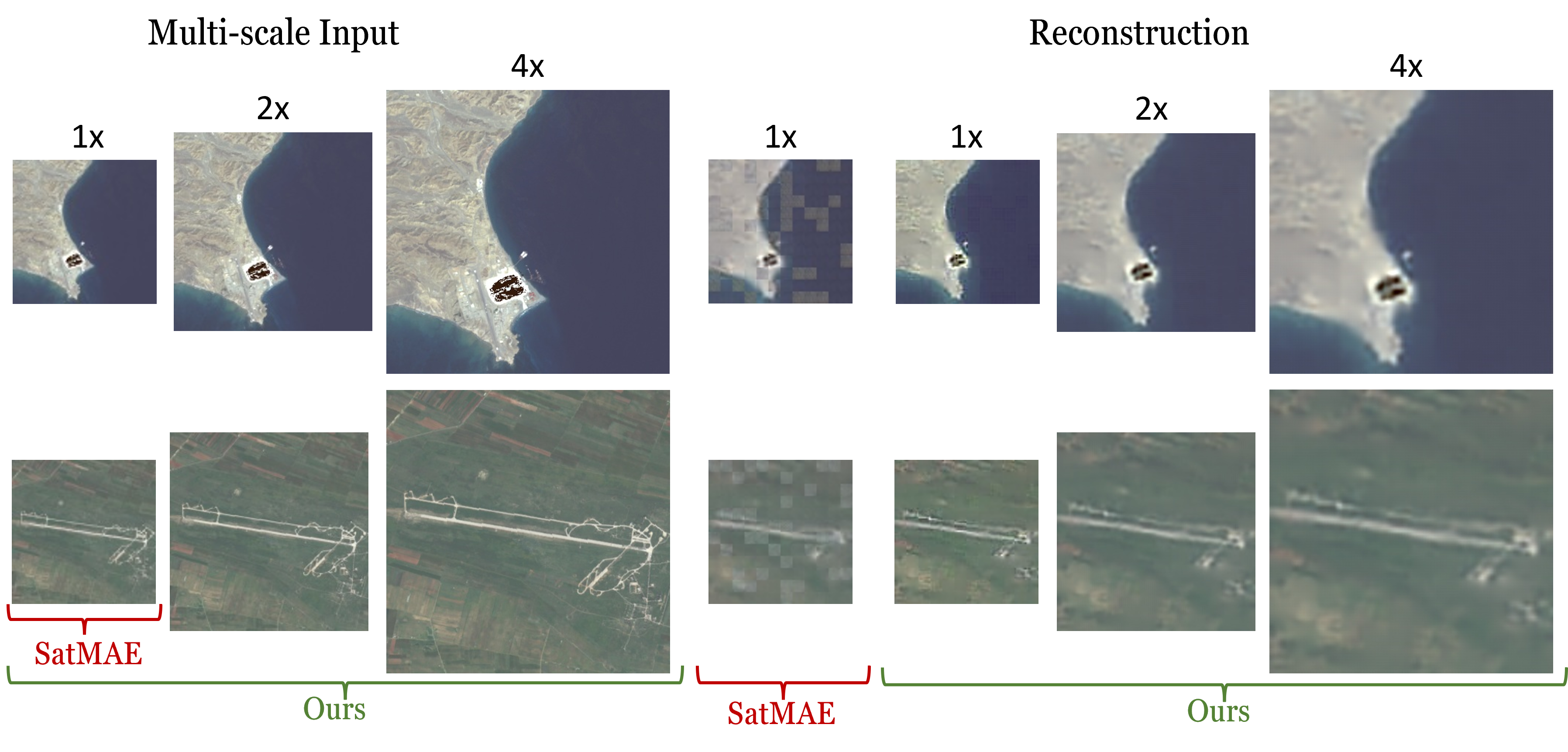}
    \caption{SatMAE++ reconstruction results at multi-scale level. Examples from fMoW-Sentinel dataset are shown here. For illustration, we show the RGB channels of the multi-spectral data here. The images are reconstructed at resolutions of $(H,W)$, $(2H,2W)$, and $(4H,4W)$, respectively. We observe that the proposed model provide better reconstruction results compared to SatMAE at resolution of $(H,W)$.}  
   
   \label{fig:reconstruction_results}
   \vspace{-0.5em}
\end{figure*}


\subsubsection{Reconstruction at Three Scale Levels}
For multi-spectral data, we reconstruct the image at three scale levels as the resolution of the model input is smaller compared to the RGB data (RGB uses $224 \times 224$ pixels). Here, we take image $\bar I$ at a higher resolution of $\mathbb{R}^{C \times 4H \times 4W}$. We then down sample the image $\bar I$ twice to obtain the images $\hat I \in \mathbb{R}^{C \times 2H \times 2W}$ and $I \in \mathbb{R}^{C \times H \times W}$, respectively. As discussed in the previous section, we reconstruct the image $F$ at spatial resolution of $(H, W)$. Afterwards, a linear projection layer is applied to project the data to feature space and input it to the upsample block. The upsample block uses the transpose convolution to increase the spatial resolution by two times which is then fed to the residual convolution block to obtain features $\hat F$ having spatial resolution $(2H, 2W)$. The features $\hat F$
are input to another upsample block to obtain features $\bar F$ with dimensions $(4H, 4W)$.
Both features $\hat F$ and $\bar F$ are projected back to the image space and L1 loss is applied to measure reconstructed features quality. The overall loss is the weighted average of the three losses given as:
\begin{equation}
\resizebox{.5\hsize}{!}{
\label{eq:losses_three_scales}
$
\begin{aligned}
    L_3 &= \frac{1}{n}\sum_{i=1}^{n}{|\bar F - \bar I|} \\
    Loss &= \alpha_1 L_1+\alpha_2 L_2+\alpha_3 L_3
\end{aligned}
$
}
\end{equation}

\section{Experiments}
In this section, we first discuss about the mainstream datasets. Then, we explain the pre-training and finetuning procedures utilized for the mainstream benchmarks. Afterwards, we discuss about the datasets used for transfer learning experiments and its finetuning methodology. We further discuss the performance of learned representations through simplified multi-scale MAE pre-training for SatMAE++ and also present the fine tuning results on downstream tasks.

\subsection{Pre-training Datasets}
We utilized two publicly available large scale datasets for pre-training the vision transformer on multi-spectral and RGB satellite data.

\noindent\textbf{fMoW-RGB:} Functional Map of the World (fMoW) \cite{christie2018fmowrgb} is a large scale publicly available dataset of high resolution satellite images. The dataset is divided into 62 categories for classification task and comprises of about 363k training and 53k test images.

\noindent\textbf{fMoW-Sentinel:} SatMAE \cite{cong2023satmae} refines and extended the fMoW-RGB for classification task and includes the Sentinel-2 data for the images. Similar to the fMoW-RGB, this dataset has 62 class categories. The dataset contains greater number of images and comprises of 712874 training, 84939 validation, and 84966 test images.

\begin{figure}[t!]
   \centering
    \includegraphics[width=\linewidth]{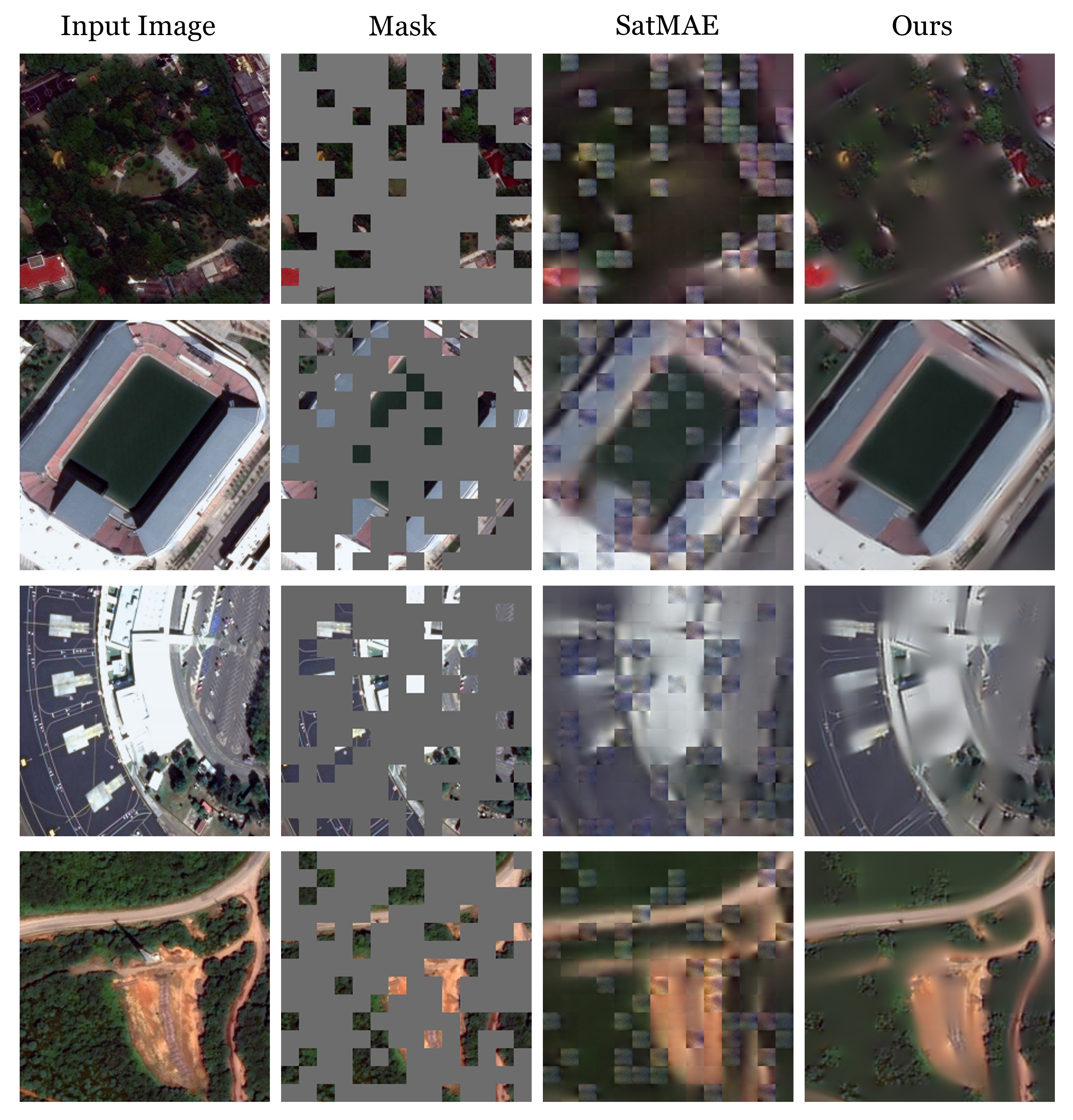}
    \caption{Here, we compare the reconstruction performance of our framework with the baseline SatMAE. We observe that the reconstruction results of SatMAE on visible patches is worse compared to the masked patches. Whereas our framework provides much better results on all the patches including the visible patches. The above reported results demonstrate the effectiveness of multi-scale pre-training framework SatMAE++.}
   \label{fig:reconstruction_comparison}
\end{figure}

\noindent\textbf{Reconstruction Results:} We present the multi-scale reconstruction results on fMoW-Sentinel dataset in Fig.~\ref{fig:reconstruction_results}. We observe the improvement in reconstruction quality of the image by employing multi-scale pre-training. Furthermore, we compare the reconstruction results of our approach with the baseline approach (SatMAE) in Fig.~\ref{fig:reconstruction_comparison}. The reconstruction quality of SatMAE on visible patches is much worse in comparison with the masked patches. However, our multi-scale approach improves the quality of reconstruction for both visible and masked patches. While achieving favorable reconstruction compared to SatMAE (see Fig.~\ref{fig:reconstruction_results} and \ref{fig:reconstruction_comparison}), we observe our approach to struggle in some cases to reconstruct fine-grained structural details.

\begin{table}[t]
\centering
\caption{ Comparison of state-of-the-art methods on the validation set of fMoW-RGB dataset. * represents that we reproduce the results using the public codebase provided by the authors and pre-train the model for 800 epochs on fMoW-RGB, which are consistent to the performance reported in ScaleMAE \cite{reed2023scalemae}}
\setlength{\tabcolsep}{8pt}
\label{tbl:finetuning_fmow_rgb}

\begin{tabular}{l|c|c}
\hline
\rowcolor{Gray_blue}
\textbf{Method} & \textbf{Backbone} & \textbf{Top1 Acc.} \\
\hline
GASSL \cite{ayush2022geographyaware} & ResNet-50 & 71.55  \\
MoCo-V2 \cite{he2020momentum} & ResNet-50 &  64.34 \\
MAE \cite{he2021maskedautoencoder} & ViT-Large & 68.4  \\
ConvMAE \cite{gao2022convmae} & Conv ViT-Large &  74.1  \\
ScaleMAE \cite{reed2023scalemae} & ViT-Large  & 77.9  \\
SatMAE \cite{cong2023satmae} & ViT-Large  & 77.84 \\
SatMAE$^*$ \cite{cong2023satmae} & ViT-Large  & 73.87 \\

\hline
\rowcolor{Gray}
\textbf{SatMAE++ (Ours)} & ViT-Large  & \textbf{78.14}  \\
\hline
\end{tabular}
\end{table}

\subsection{Pre-training and Finetuning on fMoW-RGB }
\noindent\textbf{Pre-training:} Similar to the SatMAE \cite{cong2023satmae} and ScaleMAE \cite{reed2023scalemae}, we pre-train the ViT-Large \cite{dosovitskiy2021vit} model on fMoW-RGB dataset. The configuration of model is same as utilized in \cite{cong2023satmae} i.e., input image of spatial resolution $(224 \times 224)$ and patch size of 16. During pre-training, we resize the shorter side of the image to $448$ pixels and then perform random crop of $448 \times 448$ pixels from the resized image. The image of size $448 \times 448$ pixels is then downsampled to a lower resolution of $224 \times 224$ pixels by using bilinear interpolation. We then input the image of size $224 \times 224$ pixels to the model. We use AdamW optimizer and cosine learning rate scheduler in our pre-training experiments. The initial learning rate is set to 7e-4 and batch size of 64 is used for single GPU. Similar to \cite{cong2023satmae} 75\% of the patches are masked while pre-training the model. We utilize 8 NVIDIA V100 GPUs to train the model for 800 epochs.

\noindent\textbf{Finetuning:} We finetune the ViT-Large model by loading the pre-training weights in an end-to-end manner. The initial learning rate is set to 1e-3 and batch size of 8 is used for single GPU. We use the AdamW optimizer with cosine scheduler and keep the configurations of augmentations and weight decay similar to the \cite{cong2023satmae}. We finetune the model for 50 epochs on 8 NVIDIA V100 GPUs.

\noindent\textbf{Discussion: } We present the finetuning results of our approach in Tab.~\ref{tbl:finetuning_fmow_rgb}. The state-of-the-art ScaleMAE \cite{reed2023scalemae} shows an improved performance of 77.9\% over the SatMAE \cite{cong2023satmae} by utilizing sophisticated GSD based positional encodings and Laplacian decoder. However, our approach outperforms the state-of-the-art ScaleMAE \cite{reed2023scalemae} by providing the score of 78.14\% without using the sophisticated GSD based positional encodings. Compared to the baseline method \cite{cong2023satmae}, our approach provides an absolute gain of 4.27\% over the reproduced results and 0.3\% over the numbers reported in the paper.

\begin{figure}[t!]
   \centering
    \includegraphics[width=\linewidth]{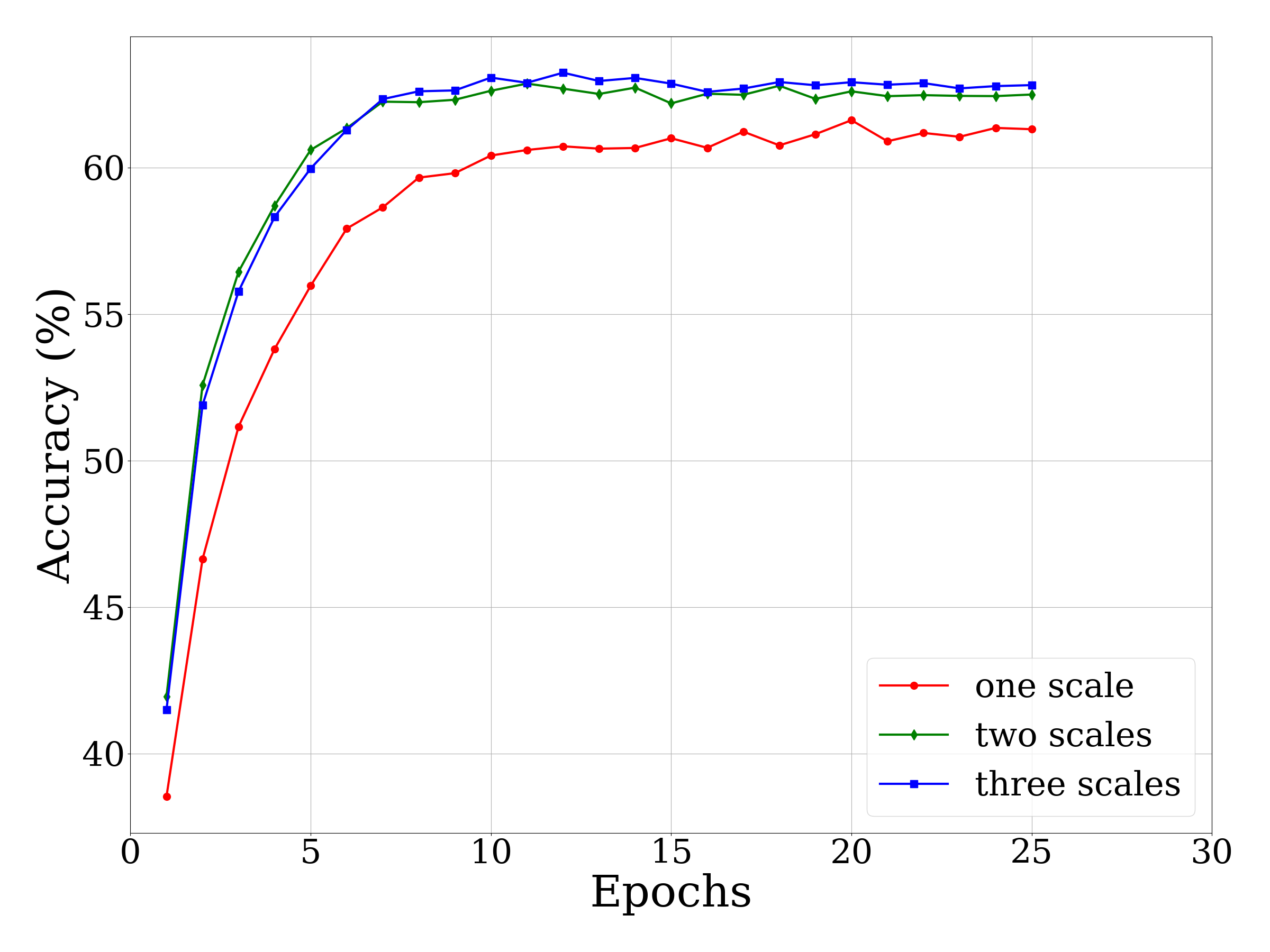}
    \caption{Illustration of finetuning convergence on validation set of fMoW-Sentinel dataset. We observe that the model pre-trained with multi-scales achieves faster convergence as compared the model pre-trained with single or less scales. The model trained with single scale achieves highest score of 61.61 at 20th epoch whereas the model that utilised three scales in pre-training converges earlier and achieves highest score at 12th epoch.}    
   \label{fig:acc_vs_epoch_fmow_sentinel}
\end{figure}


\subsection{Pre-training and Finetuning on fMoW-Sentinel }
\noindent\textbf{Pre-training:} Following \cite{cong2023satmae}, we pre-train the ViT-Large model on fMoW-Sentinel dataset having image size of $96 \times 96$ and patch size of 8. We employ the SatMAE+Group+IM strategy of grouping the channels and independent masking for the muli-spectral data. All 13 bands of the Sentinel-2 images are not utilized and we discard the channels B1, B9, and B10 during pre-training and finetuning. We follow the channel grouping approach of \cite{cong2023satmae} and create (i) group containing B2, B3, B4, B8 channels (ii) group comprising of B5, B6, B7, B8A channels, and (iii) group of B11, B12 channels, respectively. The groups are selected to have same GSD resolution. During pre-training stage, the shorter side of the image is resized to $384$ pixels and random cropping is used to obtain image of size $384 \times 384$ pixels. We then interpolate $384 \times 384$ sized image to obtain two down sampled images of sizes $192 \times 192$ and $96 \times 96$ pixels, respectively. The resized image having spatial resolution of $96 \times 96$ is input to the model. The model is pre-trained with a base learning rate of 1e-4 and batch size of 8 for 50 epochs on 8 NVIDIA V100 GPUs. The remaining settings of optimizer, learning rate scheduler, weight decay, masking ratio, are kept same as listed in \cite{cong2023satmae}.

\begin{table}[t]
\centering
\caption{ Finetuning results on the validation set of fMoW-Sentinel dataset. * represents that we reproduce the results using the publicly available codebase provided by the authors and pre-train the model on fMoW-Sentinel for 50 epochs.}
\setlength{\tabcolsep}{10pt}
\label{tbl:finetuning_fmow_sentinel}

\begin{tabular}{l|c|c}
\hline
\rowcolor{Gray_blue}
\textbf{Method} & \textbf{Backbone} & \textbf{Top1 Acc.} \\
\hline
MoCo-V3 & ViT-Base &  50.45 \\
MoCo-V3+Group & ViT-Base  & 51.33  \\
SatMAE$^*$ \cite{cong2023satmae} & ViT-Base  & 60.01 \\

\hline
\rowcolor{Gray}
\textbf{SatMAE++ (Ours)} & ViT-Base  & \textbf{62.69}  \\
\hline
ScaleMAE$^*$ \cite{reed2023scalemae} & ViT-Large  & 42.21  \\
SatMAE \cite{cong2023satmae} & ViT-Large  & 61.48 \\
SatMAE$^*$ \cite{cong2023satmae} & ViT-Large  & 61.61 \\

\hline
\rowcolor{Gray}
\textbf{SatMAE++ (Ours)} & ViT-Large  & \textbf{63.23}  \\
\hline
\end{tabular}
\end{table}

\noindent\textbf{Finetuning:} In the finetuning stage, we load the pre-training weights of the ViT-Large model and finetune it for 30 epochs. We used the base learning rate of 2e-4, input size of $96 \times 96$ pixels and patch size of 8. Following \cite{cong2023satmae}, we utilize AdamW optimizer, cosine scheduler and same data augmentations. We employ 8 NVIDIA V100 GPUs for finetuning the model.

\noindent\textbf{Discussion:} We report the state-of-the-art comparison on fMoW-Sentinel dataset in Tab.~\ref{tbl:finetuning_fmow_sentinel}. Our approach that employs multi-scale pre-training provides significant improvement over the SatMAE \cite{cong2023satmae}. In comparison to the state-of-the-art method SatMAE that utilizes ViT-Large backbone, our approach provides an improved accuracy score of 1.75\%. In case of ViT-Base backbone, SatMAE++ achieves a gain of 2.68\% in the terms of top1 accuracy compared to the SatMAE method which is even better than the score of SatMAE method obtained by utilizing the ViT-Large backbone.
We observe that the muli-scale pre-training encourage the model to learn better feature representations especially when the image resolution varies considerably in the given dataset.

\noindent\textbf{Ablation Studies:} To demonstrate the efficacy of multi-scale pre-training, we pre-train the ViT-Large model with one, two and three scales. Afterwards, we finetune the model by loading the respective pre-trained weights on fMoW-Sentinel dataset. Tab.~\ref{tbl:ablation_on_fmow_sentinel} shows the results of finetuning by loading the pre-train weights of one, two and three scales. We observe consistent improvement in the performance of the model as pre-training scales are increased. Compared to the single scale level, the improvement of the model is significant when two scales are utilized i.e., a gain of 1.25\% is achieved. However, the performance gain is 0.37\% when we shift from the two scale to three scale levels. We hypothesize that after a certain number of scale levels, the effect of multi-scale pre-training may be insignificant.

\noindent\textbf{Convergence Rate: } Fig.~\ref{fig:acc_vs_epoch_fmow_sentinel} shows the convergence of the model pre-trained with multiple scales on fMoW-Sentinel dataset. We observe that the model trained with single scale reaches its highest performance score of 61.61 at 20th epoch. Whereas when multi-scale pre-training is utilized, model converges earlier and achieves it highest performance score of 63.23 on the 12th epoch. From this, we infer that the multi-scale pre-training can uplift the performance of the models and provide faster convergence especially when data has diverse range of scale variations.

\begin{table}[t]
\centering
\caption{Ablation study demonstrating the effect of using multiple scales in pre-training. Our SatMAE++ with three scales provides a superior performance with top-1 accuracy of 63.23\%.
}
\label{tbl:ablation_on_fmow_sentinel}
\scalebox{0.9}{
    \begin{tabular}{l|c|c|c}
    \hline
    \rowcolor{Gray_blue}
    \textbf{Method} & \textbf{Backbone} & \textbf{Pre-training Scales} & \textbf{Top1 Acc.} \\
    \hline
    SatMAE & ViT-Large & 1 & 61.61  \\
    SatMAE++ & ViT-Large & 2 & 62.86  \\
    SatMAE++ & ViT-Large & 3 & \textbf{63.23}  \\
    \hline
    \end{tabular}
}
\end{table}


\subsection{Downstream Datasets} To demonstrate the effectiveness of our pre-training approach, we utilize following datasets corresponding to the land cover and multi-label classification tasks.

\noindent\textbf{EuroSAT} \cite{helber2019eurosat} is a publicly available remote sensing dataset for land use and land cover (LULC) classification. It is categorized into 10 classes and comprises of 27000 images. The dataset is available in both RGB and multi-spectral (Sentinel-2) format. Following \cite{cong2023satmae}, we use the data split provided by \cite{neumann2019indomain}.

\noindent\textbf{RESISC-45} \cite{Cheng2017resisc45} is another public remote sensing scene classification dataset comprising of 31500 images having 45 scene classes. We follow the \cite{reed2023scalemae} for train / val splits of the dataset.

\noindent\textbf{UC-Merced} \cite{Yang2010ucmerced} is a public land use remote sensing image dataset which contains 21 scene classes. Each class comprises of 100 images that are manually selected from the US regions. We follow the data split provided by \cite{neumann2019indomain} in our experiments.

\noindent\textbf{BigEarthNet} \cite{Sumbul2019bigearthnet} is a multi-label land cover classification dataset and is publicly available for research purpose. The dataset is composed of 590326 Sentinel-2 images and categorized into 19 classes. Following \cite{cong2023satmae}, we use 10\% of the train data in our experiments and utilize the data splits available at \cite{neumann2019indomain}.

\subsection{Transfer Learning on Downstream Datasets}
We finetune the pre-trained ViT-Large model (SatMAE++) on various downstream remote sensing tasks such as land cover and multi-label classification. SatMAE++ provides consistent improvement on the downstream tasks compared to the other state-of-the-art methods existing in literature.

\subsubsection{Land Cover Classification } We present the transfer learning experiments for land cover classification task on three publicly available remote sensing datasets including EuroSAT \cite{helber2019eurosat}, RESISC-45 \cite{Cheng2017resisc45}, and UC-Merced \cite{Yang2010ucmerced}. We utilize the configuration settings used in fMoW-RGB finetuning experiment for transfer learning on these datasets.

\noindent\textbf{Finetuning on EuroSAT: } We present the finetuning results on EuroSAT dataset in Tab.~\ref{tbl:finetuning_eurosat}. We observe that the multi-scale pre-training approach provides reasonable improvement over other approaches. It is notable that SatMAE++ surpasses the performance score of the state-of-the-art SatMAE without using the multi-spectral information and achieves an accuracy score of 99.01\%.

\noindent\textbf{Finetuning on RESISC-45: } Tab.~\ref{tbl:finetuning_resisc} shows the scene classification performance on RESISC-45 dataset. Among recent methods, ScaleMAE \cite{reed2023scalemae} achieves superior performance, however, ViT-Large when finetuned by using the pre-trained weights of our approach, surpasses the ScaleMAE score and achieves an accuracy of 97.48\%.

\noindent\textbf{Finetuning on UC-Merced: } We further report the effectiveness of multi-scale MAE pre-training on another popular LULC dataset (Tab.~\ref{tbl:finetuning_ucmerced}). Here, we report the results of finetuning the ViT-Large model on UC-Merced dataset by loading the pre-trained weights of SatMAE and our approach, respectively. The ViT-Large model provides the accuracy score of 94.05\% when utilizes the pre-trained weights of SatMAE. However, the performance of ViT-Large improves by 3.6\% approx when finetuning is performed by loading the pre-trained weights of our approach.

\begin{table}[t]
\centering
\caption{Land cover classification on EuroSAT dataset. $\dagger$ denotes that the model uses multi-spectral data.}
\setlength{\tabcolsep}{10pt}
\label{tbl:finetuning_eurosat}

\begin{tabular}{l|c|c}
\hline
\rowcolor{Gray_blue}
\textbf{Method} & \textbf{Backbone} & \textbf{Top1 Acc.} \\
\hline
GASSL \cite{ayush2022geographyaware} & ResNet-18 & 89.51  \\
SeCo \cite{mañas2021seco} & ResNet-18 & 93.14  \\
SatMAE \cite{cong2023satmae} & ViT-Large  & 95.74 \\
SatMAE$^\dagger$ \cite{cong2023satmae} & ViT-Large  & 98.98 \\
\hline
\rowcolor{Gray}
\textbf{SatMAE++ (Ours)} & ViT-Large  & \textbf{99.04}  \\
\hline
\end{tabular}
\end{table}

\begin{table}[t]
\centering
\caption{Finetuning results of land cover classification on RESISC-45 dataset. }
\setlength{\tabcolsep}{10pt}
\label{tbl:finetuning_resisc}

\begin{tabular}{l|c|c}
\hline
\rowcolor{Gray_blue}
\textbf{Method} & \textbf{Backbone} & \textbf{Top1 Acc.} \\
\hline
MAE \cite{he2021maskedautoencoder} & ViT-Large  & 93.3 \\
ConvMAE \cite{gao2022convmae} & ViT-Large  & 95.0 \\
SatMAE \cite{cong2023satmae} & ViT-Large  & 94.8 \\
ScaleMAE \cite{reed2023scalemae} & ViT-Large  & 95.7 \\
\hline
\rowcolor{Gray}
\textbf{SatMAE++ (Ours)} & ViT-Large  & \textbf{97.48}  \\
\hline
\end{tabular}
\end{table}

\begin{table}[t]
\centering
\caption{Finetuning results of land cover classification on UC-Merced dataset. }
\setlength{\tabcolsep}{10pt}
\label{tbl:finetuning_ucmerced}

\begin{tabular}{l|c|c}
\hline
\rowcolor{Gray_blue}
\textbf{Method} & \textbf{Backbone} & \textbf{Top1 Acc.} \\
\hline
SatMAE \cite{cong2023satmae} & ViT-Large  & 94.05 \\
\textbf{SatMAE++ (Ours)} & ViT-Large  & \textbf{97.62}  \\
\hline
\end{tabular}
\end{table}

\begin{table}[t]
\centering
\caption{Multi-Label classification results on BigEarthNet \cite{Sumbul2019bigearthnet} dataset.
Following \cite{cong2023satmae}, we use mean Average Precision (mAP) metric and newer set of class labels. The reported results utilize 10\% of training data.
$\ddagger$ denotes that model uses RGB bands only.
}
\setlength{\tabcolsep}{12pt}
\label{tbl:finetuning_bigearthnet}

\begin{tabular}{l|c|c}
\hline
\rowcolor{Gray_blue}
\textbf{Method} & \textbf{Backbone} & \textbf{mAP} \\
\hline
GASSL \cite{ayush2022geographyaware} & Resnet-50  & 80.20 \\
SeCo$^\ddagger$ \cite{mañas2021seco} & Resnet-50  & 82.62 \\
SatMAE \cite{cong2023satmae} & ViT-Large  & 82.13 \\
\textbf{SatMAE++ (Ours)} & ViT-Large  & \textbf{85.11}  \\
\hline
\end{tabular}
\end{table}

\subsubsection{Multi-Label Classification} 
Finally, we report the performance of our approach on multi-label classification task. We finetune the ViT-Large model on BigEarthNet \cite{Sumbul2019bigearthnet} dataset by loading the pre-train weights of fMoW-Sentinel dataset. We keep the same configurations for finetuning as utilized in fMoW-Sentinel experiment. As the task is multi-label classification, therefore \textit{soft target cross entropy} loss is replaced with the \textit{multi-label soft margin} loss. Following \cite{cong2023satmae}, we use the average precision metric as a performance measure of the model.

Tab.~\ref{tbl:finetuning_bigearthnet} shows the finetuning results on BigEarthNet dataset. Similar to the other downstream tasks, our framework performs favorably in the case of multi-label classification task. Compared to the state-of-the-art frameworks, SeCo \cite{mañas2021seco} performs fairly well and achieves the state-of-the-art score of 82.62\%. SatMAE \cite{cong2023satmae} has a slightly lower score of 82.13\% as compared to the SeCo \cite{mañas2021seco}. However, our framework provides significant improvement and achieves an average precision score of 85.11\%.

\section{Conclusion}
Remote sensing imagery offers a wide range of resolutions and spectral bands having multi-scale information incorporated within it. Existing state-of-the-art methods struggle to effectively utilize the multi-scale information along with the multi-spectral data.
We propose a framework, named SatMAE++,
to incorporate multi-scale information thereby improving the model performance and achieving faster convergence during finetuning. Our SatMAE++ is easily extensible to multiple scale levels and is not restricted to single type of data modality. Extensive experimentation on several downstream and mainstream datasets reveal the efficacy of our approach. Future work includes extending the proposed multi-scale pre-training for dense prediction tasks.

\bibliographystyle{ieeenat_fullname}
\bibliography{ref}

\end{document}

%% file: preamble.tex
\usepackage[dvipsnames]{xcolor}
